# Memory Retrieved From Single Neurons

Subhash Kak

**Abstract:** The paper examines the problem of accessing a vector memory from a single neuron in a Hebbian neural network. It begins with the review of the author's earlier method, which is different from the Hopfield model in that it recruits neighboring neurons by spreading activity, making it possible for single or groups of neurons to become associated with vector memories. Some open issues associated with this approach are outlined. It is suggested that fragments that generate stored memories could be associated with single neurons through local spreading activity.

## Introduction

How is biological memory organized? In patterns of activity as assumed by artificial neural network models (e.g.[1]-[6]) or as information stored in specific location as in some recent experimental findings [7],[8]? Is the neuron function low level so that each neuron processes small amount of information or is it high level so that it can deal with higher level abstract categories? The objective of this paper is to provide a specific mathematical view on these questions that is based on the feedback artificial neural network model. But by way of motivation, we will speak of recent results from neuroscience.

The memory capacity of the neurons for assumed patterns of activity is about equal to the number of neurons if they are binary [2],[3],[9]-[11], and substantially higher if they are non-binary [12]. If memory is stored in specific locations, the capacity would be of similar amount.

Learning requires networks to have the capacity to generalize very quickly. The motivation to model short-term memory led to the development of the corner classification (CC) family of the instantaneously trained neural networks (ITNN) [4]-[6] that learn instantaneously and have good generalization performance. But as part of a more versatile memory, one would expect such memories to have strong feedback connections with other processing modules.

There is evidence in support of visual representation that is invariant to metrical properties in the primate brain by single neurons. Quiroga *et al* have shown [8] that neurons in the human medial temporal lobe (MTL) "are selectively activated by strikingly different pictures of given individuals, landmarks or objects and in some cases even by letter strings with their names." One would expect that the visual system of other higher animals has similar characteristics.

According to Lin, Osan and Tsien [13], network-level coding units, or neural cliques, "are functionally organized in a categorical and hierarchical manner, suggesting that internal representations of external events in the brain is achieved not by recording exact details of those events, but rather by recreating its own selective pictures based on cognitive importance…. In addition, activation patterns of the neural clique assemblies can be converted to strings of binary codes that would permit universal categorizations of internal brain representations across individuals and species." Further characteristics of Hebbian learning are provided in [14] and [15].

According to Barlow [16] and others, one must invoke higher level cognitive neurons. On the other hand, Softky [17] sees the experimental data to be consistent with both the "simple" model, in which each neuron performs a simple computation and transmits a small amount of information, and the competing "efficient" model, in which a neuron transmits large amounts of information through precise, complex, single-spike computations. Thompson [18] finds the basic memory trace to be localized in the cerebellum, and putative higher-order memory traces to be characterized in the hippocampus.

Memories may be viewed from the perspective of artificial intelligence or natural context [19]-[23], and here one may also speak of recent investigations that consider the possible role of quantum theory [24]-[26]. Another angle, from the perspective of mathematical models of neural networks, is that of indexing of memories [27]-[29]. This seems to be particularly relevant in the ability of individuals to recall a tune from a note, and the fact that we can recognize object and individuals even when seen in strikingly different situations.

It should be noted that while the Hopfield (or feedback) neural network model [2],[3], which may be viewed as a generalization of the idea of storage in terms of eigenvectors for a matrix, is a model for storage, it is not a model for recall by index. In this model we can only check if a given memory is stored, but since the memory is not localized it cannot be recalled by index. This is a lacuna of the model that we wish to address by examining further a method, proposed earlier by the author, where the activity spreads locally.

The brain is composed of several modules each of which is essentially an autonomous neural network. Thus the visual network responds to visual stimulation and also during visual imagery, which is when one sees with the mind's eye. Likewise, the motor network produces movement and it is active during imagined movements. However, although the brain is modular, a part of it, located for most people in the left hemisphere, monitors the modules and interprets their individual actions in order to create a unified idea of the self. In other words, there is a higher integrative or interpretive module that synthesizes the actions of the lower modules [19].

In this paper, we first review our earlier method, applicable to feedback networks, of recruitment of neighboring neurons by spreading activity, in which unique fragments become associated with vector memories [27]. Next, we show how local updating may be performed to obtain fragments from single neurons. We propose that a combination of these two methods will be an effective recall system. We speak of some open issues associated with this approach to neural memories.

**The Generator Model**

The neural network is trained by Hebbian learning, in which the connectivity (synaptic strength) of neurons that fire together strengthens and that of those that don't gets weakened. The interconnection matrix $T = \sum x^{(i)} x^{(i)t}$, where the memories are column vectors, $x^i$, and the diagonal terms are taken to be zero.

A memory is stored if

$$x^i = \text{sgn}(Tx^i) \qquad (1)$$

where the *sgn* function is 1 if the input is equal or greater than zero and -1, if the input is less than 0.

In the generator model [27], the memories are recalled by the use of the lower triangular matrix $B$, where $T = B + B^t$. Effectively, the activity starts from one single neuron and then spreads to additional neurons as determined by $B$ (Figure 1). Note that with each update, the fragment enlarges by one neuron and it is fed back into the circuit.

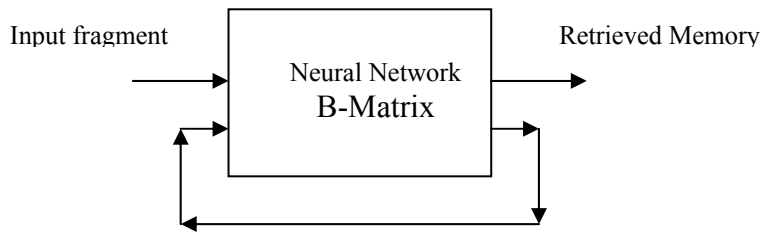

**Figure 1.** Obtaining memory from its fragment

The spreading function where the information spreads from neuron 1 to neuron 2 and so on has an implicit assumption regarding the geometrical (proximity) relationship amongst the neurons.

Starting with the fragment $f^i$, the updating proceeds as:

$$f^i \text{ (new)} = \text{sgn}(Bf^i \text{ (old)}) \qquad (2)$$

where the original fragment values are *left unchanged* on the neurons and the updating proceeds one step at a time.

This will be illustrated by means of two examples.

**Example 1.** Consider Hebbian learning applied to the following three memories, $x^i$, $i=1,2,3$, where each memory is a column vector that is shown below as its transpose:

$$
\begin{aligned}
x^{1t} &= 1 \quad 1 \quad 1 \quad 1 \quad 1 \\
x^{2t} &= 1 \quad -1 \quad -1 \quad -1 \quad 1 \\
x^{3t} &= 1 \quad 1 \quad -1 \quad -1 \quad -1
\end{aligned}
\qquad (3)
$$

It may be easily checked that all the three memories are indeed stored for this example.

$$
T = \begin{bmatrix} 0 & 1 & -1 & -1 & 1 \\ 1 & 0 & 1 & 1 & -1 \\ -1 & 1 & 0 & 3 & 1 \\ -1 & 1 & 3 & 0 & 1 \\ 1 & -1 & 1 & 1 & 0 \end{bmatrix} \qquad (4)
$$

The dynamics of the spreading function will be governed by the triangular matrix $B$:

$$
B = \begin{bmatrix} 0 & 0 & 0 & 0 & 0 \\ 1 & 0 & 0 & 0 & 0 \\ -1 & 1 & 0 & 0 & 0 \\ -1 & 1 & 3 & 0 & 0 \\ 1 & -1 & 1 & 1 & 0 \end{bmatrix} \qquad (5)
$$

Let the first fragment be the bit "1", which is clamped on neuron number 1, and, therefore, in the updating below, this value will not change.

$$
f^1(new) = \operatorname{sgn} \begin{bmatrix} 0 & 0 & 0 & 0 & 0 \\ 1 & 0 & 0 & 0 & 0 \\ -1 & 1 & 0 & 0 & 0 \\ -1 & 1 & 3 & 0 & 0 \\ 1 & -1 & 1 & 1 & 0 \end{bmatrix} \begin{bmatrix} 1 \\ 0 \\ 0 \\ 0 \\ 0 \end{bmatrix} = \begin{bmatrix} 1 \\ 1 \\ 0 \\ 0 \\ 0 \end{bmatrix} \Longrightarrow \begin{bmatrix} 1 \\ 1 \\ 1 \\ 1 \\ 1 \end{bmatrix} \qquad (6)
$$

At the second pass, $f^1\ (new)= [1\ 1\ 1\ 0\ 0]^t$. The operation in (6) proceeded on a neuron-by-neuron basis. The fragment increased from "1" to "1 1" and it was fed back into the circuit and increased to "1 1 1" and so on until memory #1 is retrieved.

If the starting fragment is "1 -1", we get the following:

$$f^2(new) = \text{sgn} \begin{bmatrix} 0 & 0 & 0 & 0 & 0 \\ 1 & 0 & 0 & 0 & 0 \\ -1 & 1 & 0 & 0 & 0 \\ -1 & 1 & 3 & 0 & 0 \\ 1 & -1 & 1 & 1 & 0 \end{bmatrix} \begin{bmatrix} 1 \\ -1 \\ 0 \\ 0 \\ 0 \end{bmatrix} \Rightarrow \begin{bmatrix} 1 \\ -1 \\ -1 \\ -1 \\ 1 \end{bmatrix} \qquad (7)$$

For the fragment is "1 1 -1", we get:

$$f^3(new) = \text{sgn} \begin{bmatrix} 0 & 0 & 0 & 0 & 0 \\ 1 & 0 & 0 & 0 & 0 \\ -1 & 1 & 0 & 0 & 0 \\ -1 & 1 & 3 & 0 & 0 \\ 1 & -1 & 1 & 1 & 0 \end{bmatrix} \begin{bmatrix} 1 \\ 1 \\ -1 \\ 0 \\ 0 \end{bmatrix} \Rightarrow \begin{bmatrix} 1 \\ 1 \\ -1 \\ -1 \\ -1 \end{bmatrix} \qquad (8)$$

In other words, if the 5 neurons were labeled as 1,2,3,4,5, then the following fragments (shown within parentheses) suffice to retrieve the three memories.

    (1)2345, (12)345, (123)45                                                (9)

The unique smallest fragment for the first memory is "1", for the second one it is "1 -1", and for the third one it is "1 1 -1". The first memory is retrieved by the first neuron only; the second requires the specification of the first two neuron values; and the thirds requires the specification of the first three neuron values. (The fact that the fragments are 1, 2, and 3 bits long is a coincidence.) These memory fragments that lead to the recall of the correct memory were called generators. Clearly, complement as well as spurious memories will have their own generators.

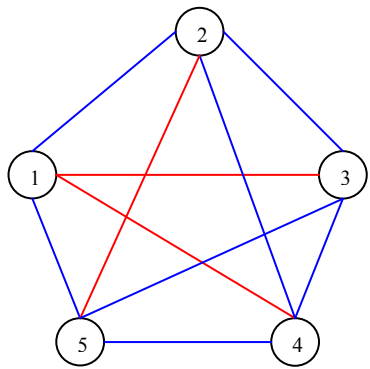

**Figure 2.** Interconnection graph for five neurons of Example 1
(blue represents positive correlation and red negative correlation)

**Generation of Initial Fragments**

The previous model of recall depended on the cumulative effect of several neurons in the spreading of the activity. In certain cases one could generate fragments from local activity that spreads from single neurons. *It spreads to other neurons based on correlation relationship, strength of interaction, and other factors.* In some cases the spreading will be decided by the physical proximity of other neurons, and in others it might be additional information coded, for example, in the arrival rates of the spikes.

**Example 1 (contd).** The interconnection graph for Example 1 is shown in Figure 2, where blue and red represent positive and negative correlations, respectively. This interconnection graph does not take into account the strengths of the connections. This graph will help determine how activations spread locally.

We have already seen that the *B* matrix generates memory #1 from a single neuron. But the fragment "1 -1", required to generate memory #2, cannot be generated from neuron 1 using local spreading.

On the other hand, for memory #3, neuron 1 with a 1 on it, will by local spreading, produce 1 on neuron 2 and -1 on neuron 3, which is the fragment "1 1 -1" that generates memory #3.

It is possible for a single neuron value, corresponding, for example, to the first, the fourth, and the second neurons, to be enough to retrieve each of the three memories:

$$(1)2345, \quad (4)1325, \quad (2)1345 \tag{10}$$

But this requires the specification of the update order (as side information) to generate the memories.

**Example 2.** Consider Hebbian learning applied to the three memories:

$$\begin{aligned} x^{1t} &= 1 \quad 1 \quad 1 \quad 1 \\ x^{2t} &= 1 \quad -1 \quad -1 \quad 1 \\ x^{3t} &= -1 \quad 1 \quad 1 \quad -1 \end{aligned} \tag{11}$$

In reality, there are only two unique memories in this example, since memories 2 and 3 are complements of each other. The interconnection matrix for this example is:

$$T = \begin{bmatrix} 0 & -1 & -1 & 3 \\ -1 & 0 & 3 & -1 \\ -1 & 3 & 0 & -1 \\ 3 & -1 & -1 & 0 \end{bmatrix} \tag{12}$$

This may be represented by the graph of Figure 2.

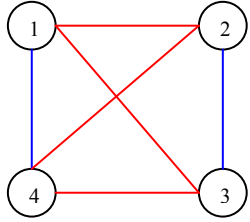

**Figure 3.** Graph for Example 2

All the three memories can be confirmed to be stored using the formula $x^i = \text{sgn}(Tx^i)$.

The B matrix for this example is:

$$B = \begin{bmatrix} 0 & 0 & 0 & 0 \\ -1 & 0 & 0 & 0 \\ -1 & 3 & 0 & 0 \\ 3 & -1 & -1 & 0 \end{bmatrix} \quad (13)$$

The fragment "1 1" clamped to the first two neurons leads to memory #1: 1 1 1 1 as seen below in the sequence $(1\ 0\ 0\ 0) \rightarrow (1\ 1\ 0\ 0) \rightarrow (1\ 1\ 1\ 0) \rightarrow (1\ 1\ 1\ 1)$:

$$\text{sgn} \begin{bmatrix} 0 & 0 & 0 & 0 \\ -1 & 0 & 0 & 0 \\ -1 & 3 & 0 & 0 \\ 3 & -1 & -1 & 0 \end{bmatrix} \begin{bmatrix} 1 \\ 1 \\ 0 \\ 0 \end{bmatrix} \Rightarrow \begin{bmatrix} 1 \\ 1 \\ 1 \\ 1 \end{bmatrix} \quad (14)$$

The fragment "1" clamped to the first neuron leads to memory #2: 1 -1 -1 1 as seen below:

$$\text{sgn} \begin{bmatrix} 0 & 0 & 0 & 0 \\ -1 & 0 & 0 & 0 \\ -1 & 3 & 0 & 0 \\ 3 & -1 & -1 & 0 \end{bmatrix} \begin{bmatrix} 1 \\ 0 \\ 0 \\ 0 \end{bmatrix} \Rightarrow \begin{bmatrix} 1 \\ -1 \\ -1 \\ 1 \end{bmatrix} \quad (15)$$

If "-1" is clamped on first neuron we obtain the complement memory #3: -1 1 1 -1.

As aside, local updating procedures beginning with single neurons lead to only the following two states (and their complements and cyclic shifts):

```
1  -1  -1   1
1  -1   1  -1
```

It may be easily confirmed that the one-bit generators for these two memories can be:

$$x^{2t} = (1)\ 2\ 3\ 4$$
$$x^{3t} = (4)\ 2\ 1\ 3$$

The virtue of the generator model is that it does not have to assume that the neurons are firing in synchrony and it does not view the updating of the neuron values to be done on a full vector basis. In a physical system, the mutual distances between the neurons will vary and the updating will proceed amongst the immediate neighbors first.

The recall procedure implies that focus on a specific neuron will trigger the memory corresponding to that neuron.

In a binary network, if each neuron is a generator of a valid memory, the maximum capacity (including complements and spurious memories) of the network will be *2N*, where *N* is the number of neurons. Due to the asymmetry of the updating process (because of the asymmetry of the *sgn* function) not all complements are valid memories.

**Problems with the Generator Model**

Since unique fragments will recall memories using the *B*-matrix, the local update mechanism needs to be used only to generate the unique fragments. Some interesting problems associated with the one-bit generator model are:

- Are there other efficient mechanisms to guide the update order for a specific memory? If the update order is primarily geometrical relationship between the neurons, not all memories would be recalled. Thus, the recalled memories from the fragments would be "colored" by the physical features of the neuron arrangements. This would be so even though, in principle, the stored memory can be verified to be so if it is presented to the neural network.

- What is the nature of the relationship between the size of the network and the number of one-bit generators?

- How many unique generators are associated with each memory? The cyclically shifted versions of each memory will be considered to be identical for this question.

- What is complexity of finding the order of updates for each generator for a specific memory if a local update procedure is used?

- Can all memories be associated with single neurons where the update order is specified in an efficient manner by the use of side information?

- How can this model be generalized to non-binary and continuous valued neurons?

Thus, several computational properties of this model need to be investigated.

**Discussion**

In the examination of learning that combines lower-level features into higher-level chunks, it has been argued that humans encode not the full correlational structure of the input, but rather higher level representations of it [30]-[39]. Other studies have shown that in face recognition, for familiar faces that are associated with relational cues additional brain regions are recruited [33]. Although the mechanism proposed in this paper does not provide a means to obtain higher-level representations, its indexing feature could, in principle, facilitate such representations.

Specifically, we examined the problem of accessing a vector memory from a single neuron in a Hebbian neural network. A method of recruitment of neighboring neurons by spreading activity was described, in which single neurons become associated with vector memories. This approach is compatible with the view that although each neuron performs low level processing it is nevertheless able to access more substantial information by virtue of its relationship with its neighbors.

Since the update order for many neurons will be determined by the proximate neurons, there would be a tendency for the memories to take stable forms that are related to this geometry. If the update mechanism has a component that reflects the geometrical relationship between the neurons, not all memories may be recalled. Thus, the recalled memories from the fragments would be "colored" by the physical features of the neuron arrangements. This would be so even though, in principle, the stored memory can be verified to be so if it is presented to the neural network. This provides a basis of the distinction between what is accessible to recall and what is accessible to verification. Memories whose generation is guided by the geometric relationship amongst neurons may explain archetype images or never witnessed images in dreams.

**Acknowledgement.** The author wishes to thank Abhinav Gautam, MD, and Arushi Kak for discussions.